\documentclass[]{fairmeta}
\usepackage{amsfonts}

\title{Populate-A-Scene: \\Affordance-Aware Human Video Generation}

\author[1,2,*]{Mengyi Shan}
\author[1]{Zecheng He}
\author[1]{Haoyu Ma}
\author[1]{Felix Juefei-Xu}
\author[1]{{Peizhao Zhang}}
\author[1]{Tingbo Hou}
\author[1]{Ching-Yao Chuang}

\affiliation[1]{GenAI at Meta}
\affiliation[2]{University of Washington}

\contribution[*]{Work done at Meta}

\abstract{Can a video generation model be repurposed as an interactive world simulator?
We explore the affordance perception potential of text-to-video models by teaching them to predict human-environment interaction. Given a scene image and a prompt describing human actions, we fine-tune the model to insert a person into the scene, while ensuring coherent behavior, appearance, harmonization, and scene affordance. Unlike prior work, we infer human affordance for video generation (i.e., where to insert a person and how they should behave) from a single scene image, without explicit conditions like bounding boxes or body poses. An in-depth study of cross-attention heatmaps demonstrates that we can uncover the inherent affordance perception of a pre-trained video model without labeled affordance datasets. }

\date{\today}
\correspondence{Mengyi Shan at \email{shanmy@cs.washington.edu}}

\begin{document}

\maketitle
\begin{figure}[t]

\centering
    \includegraphics[width=\linewidth]{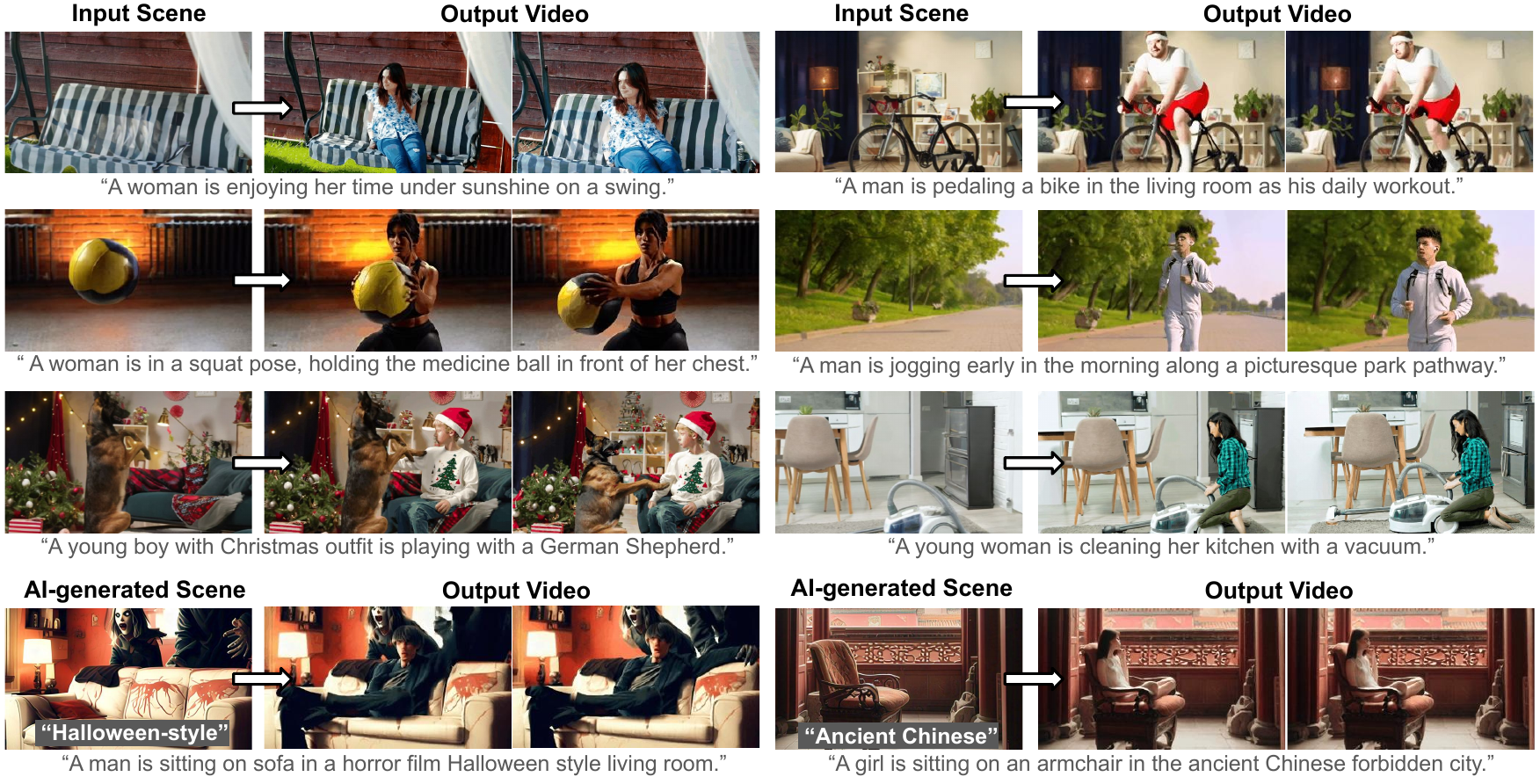}

    \captionof{figure}{We repurpose a text-to-video generation model as a human-world interaction simulator. Given a scene image and a prompt, our model inserts a person into the environment and generates a video of them naturally interacting with the scene. The scene can be real images (top) or synthesized by image generative models (bottom). Notably, there is no need for any mask, location bounding boxes, or pose sequences to guide the human insertion -- our method takes care of affordance prediction entirely within the video model. %The model infers plausible position and pose given the environment context, harmonizes the visual effect, creates human appearance as described, preserves scene details, and animates the person naturally to perform the desired action. 
    }
\label{fig:teaser}
\end{figure}

\section{Introduction}
\label{sec:intro}

Scaling data, compute, and model parameters in video generation models presents a promising avenue for developing highly capable simulators that can accurately replicate complex physical worlds \citep{meta2024moviegencastmedia, brooks2024video}, complete with diverse objects and people that interact and coexist within them. Nevertheless, humans are not merely passive observers, but rather active participants in the world. Human understanding of affordance~\cite{koffka1999principles, gibson1996senses, norman2013design} enables purposeful engagement with surroundings and adaptive behavior by recognizing potential actions afforded by an object's physical properties. It remains unclear whether video generation models can interpret and replicate intricate semantic aspects of the world, such as contextual understanding and dynamic behavior, beyond the capabilities of traditional graphics pipelines.

% Human perceives the surrounding scene based on what interactions they can make with it, where these interactive might happen, and how the environment updates as an effect. Such observation drives us to the psychological concept of affordance and signifier~\cite{koffka1999principles, gibson1996senses, norman2013design}: the former refers to the potential actions that an object allows or affords a user, based on its physical properties and the user's capabilities, and the latter marks the locations of such actions. By recognizing scene affordance and signifier, computer vision systems can predict human-object interaction, which is crucial for tasks such as robotic manipulation~\cite{lin2023mira, bharadhwaj2023robot, du2023learning}, visual navigation~\cite{kumar2018visual}, and human-object interaction~\cite{chao2015object, gkioxari2018object, shan2020hand, huang2024chatscenebridging3dscene}. 

Affordance, or ``opportunities for interaction'' \citep{gibson1996senses}, has inspired extensive research in vision and psychology. Traditional affordance prediction relies on data-driven approaches using 3D information \citep{hassan2021interaction}, specifically labeled datasets \citep{wang2017binge, gupta2011affordance, fouhey2012people, delaitre2012scene, chen2023grounding}, or one-shot large foundational models \citep{li2024oneshot}. However, these methods rely on domain-specific annotations, which are challenging to obtain. In contrast, recent advancements in generative models offer the potential to create realistic human-scene media content using vast amounts of in-the-wild media data. \citep{kulal2023affordance}, for example, predicts a human’s pose and appearance in a scene but is restricted to static images with a given position mask.

% Traditional works learn to predict affordance aided by 3D information~\cite{hassan2021interaction}, specifically-labeled datasets~\cite{wang2017binge, gupta2011affordance, fouhey2012people, delaitre2012scene, chen2023grounding}, or large foundational models~\cite{li2024oneshot}. In the 3D domain, affordance information is utilized to generation human-scene interaction based on environment geometry and semantics~\cite{li2024genzi, wang2024move}. All the classical methods concerning human-scene interaction, however, requires domain-specific annotated data which is hard to obtain. On the other hand, recent progresses in generative models shows potential to create realistic human-scene media contents with massive in-the-wild media data. One example is ~\cite{kulal2023affordance} which infers the pose and appearance of an inserted human given scene and a mask, but they only generate static images.

%We thus raise the question: Could visual generative models, like text-to-image or video models, automatically ``locate'' and ``predict'' the affordance and signifier information through content generation? We answer these questions by tuning a text-to-video model with interactive video data and illustrate its ability to populate an empty scene with diverse set of human interactions, implicitly inferring affordance and signifier. 

In this work, we demonstrate that %text-to-video models implicitly learn semantic affordance through the generative process. 
%Imagine you are a movie director tasked with creating a short film: you start by constructing a scene, populating it with objects, and then directing an actor to interact within this environment. Now, consider the role of the generation model. 
throughout the intricate process of video generation, the model learns to generate human activities and motions that adhere to the affordance constraints dictated by the physical world. To better study affordance modeling, we propose augmenting a pre-trained text-to-video model~\cite{meta2024moviegencastmedia} with an additional scene conditioning branch. This modification formulates the problem as a conditional video generation task: given a scene represented by an image, the model is tasked with introducing natural human motion and interactions to the scene. We discover that pre-trained video generation models can rapidly adapt to this new task by fine-tuning on a relatively small-scale scene conditioning dataset. We then validate the affordance perception abilities through an extensive study of the cross-attention feature heatmaps, a key module that enables the model to follow language prompts. 

% We pose our problem as a conditional video generation task given image (scene) and text (action) signal. We start with a pre-trained text-to-video model Movie Gen~\cite{meta2024moviegencastmedia}, and fine-tune it on video data containing human activities. We remove human occurrence in the starting video frame by inpainting, and rewrite the text prompt to only describe human action but not the background. Both the scene image and action prompt are used as conditions to train the video model through a grounding and fusing module inspired by conceptual grounding literature~\cite{liu2023grounding}. The scene image's latent embedding is additionally concatenated to the noisy latent to ensure background consistency. 

% The formulation pushes the model to not only learn \textit{affordance} in terms of environment properties suitable for certain interaction, but also \textit{signifiers}~\cite{norman2013design}, visual cues that indicate the presence of an \textit{affordance}. Notably, 

Unlike prior work, our model does not require input masks, bounding boxes, or pose sequences to specify regions or patterns of human behavior, which makes it an interaction \textit{simulator} that reasons about semantics and affordance properties in the scene, instead of merely a human \textit{renderer} that turns given pose signals into pixels. During inference, the model can process a wide array of environment-action combinations to generate diverse interactive videos, not limited to interaction with the single, salient object in a complicated scene. Fig. \ref{fig:teaser} demonstrates results of our model without aggressive cherry-picking. In particular, the last row of Fig. \ref{fig:teaser} illustrates a "movie studio" pipeline where input scenes are generated using a text-to-image model~\citep{dai2023emu}, and our model seamlessly integrates actors into these scenes without requiring 3D capture. Our results lower the barrier for amateur AI video creators by eliminating the need for explicit body poses signals, as they are required in most AI human video models but challenging to synthesize.

In short, this work makes the following contributions:
\begin{itemize}
    \item We address affordance-aware human video generation, where we generate video of subject(s) interacting with a given environment image, \textit{without} telling the model where the subject(s) are and how their poses look like.
    \item We apply the dual-stream conditioning mechanism with a minimal grounding module to model affordance, and thus reveal the affordance capabilities of video generation models through in-depth analysis.
    \item We demonstrate our model's ability to generalize across diverse environments and actions through a synthetic benchmark created with vision-language models.
\end{itemize}

\section{Related Works}
\label{sec:related_works}

\noindent \textbf{Text-to-video generative models. } 
Text-to-video generation aims to synthesize plausible, temporally coherent, and optionally condition-aligned video sequences from textual prompts. Recent rapid advancements in text-to-video models have been phenomenal~\cite{ho2022imagenvideohighdefinition, singer2022makeavideotexttovideogenerationtextvideo, ge2024preservecorrelationnoiseprior, blattmann2023videoldm, brooks2024videoworldsimulators, meta2024moviegencastmedia, wang2023modelscope, bartal2024lumiere, chen2024videocrafter2, esser2023structure, he2022lvdm}. Current work explores replacing the traditional U-Net with a Transformer ~\cite{vaswani2017attention} architecture~\cite{gupta2023walt, ma2024latte, brooks2024videoworldsimulators, meta2024moviegencastmedia}, inspired by the promising text-to-image generative results from DiT~\cite{peebles2022dit}. We start from a pre-trained Transformer-based text-to-video model Movie Gen~\cite{meta2024moviegencastmedia} and explore its ability to perceive affordance through minimal fine-tuning on human-scene interaction data. Some text-to-video tasks augment the model with an image as a starting frame and use prompts to describe the style or motion in the video~\cite{zeng2023makepixelsdance, gong2024atomovideo, ren2024consisti2v}. Our task differs in that we give the model an empty scene frame that is not supposed to appear in the video, but provides a ``playground'' for population. %Jiang \etal ~\cite{jiang2023videobooth} is closer to our high-level idea in using image as a prompt to video diffusion model, but they use the image prompt to specify subject identity, while we use it for reasoning about scene affordance. 

\noindent \textbf{Human video generation.}
Human video generation evolves alongside rapidly advancing generic video generative models. Generating realistic human content is inherently challenging due to complex body topology, strong priors on interaction plausibility, and audiences' sensitivity to even minor artifacts. Existing methods use motion guidance to improve video faithfulness, leveraging signals such as  OpenPose~\cite{hu2023animateanyone, wang2023disco}, DensePose~\cite{xu2023magicanimate, karras2023dreampose}, SMPL~\cite{zhu2024champ}, or a driving video~\cite{yatim2024space}. These works focus on human video generation with the subject as the sole salient element, without modeling human-environment interaction. Our work differs in that we reason about natural human-scene interaction without compromising human quality. Notably, our method requires no auxiliary conditions such as position bounding boxes~\cite{singh2023smartmask, kulal2023affordance} or motion sequences, relying instead on the internal affordance inference potential of video models.

\noindent \textbf{Human-scene interaction modeling.} 
%Previous works have attempted to insert human as an object into existing images, videos or 3D scenes. 
A fundamental task in human-environment modeling is motion prediction in 3D scenes~\cite{li2024genzi, wang2024move, kim2025david}. Related work in 2D explores interaction image and video generation from a scene, mostly using some location or body pose signals as condition~\cite{ostrek2023environmentspecificpeople, saini2024inviobjectinsertionvideos, yang2024amg, hu2025animateanyone2}. Kulal~\cite{kulal2023affordance} and Cao~\cite{cao2025disa} claim to predict affordance by inserting a human subject into a static scene, but they require a bounding box as input indicating the position. Shan~\cite{shan2023asv} insert moving humans into a street scene, but restrict actions to predefined walking motions. Singh~\cite{singh2023smartmask} predict fine-grained masks for human insertion based on scene and text descriptions but do not explicitly model human-scene interaction. Jin~\cite{jin2025unicanvas} builds on similar ideas as ours, but focus on static images with non-human objects, which in nature lack intricate interactive dynamic behaviors. Our work instead requires no semantic priors for where and how human-scene interaction occurs.

\noindent \textbf{Affordance.}
Psychologist J.J. Gibson defines \textit{affordance} as the possibilities an environment offers an individual~\cite{gibson1996senses, norman2013design} and views affordance perception as essential to socialization. Inspired by this concept from cognitive psychology, computer vision research explores scene and object affordance prediction~\cite{chuang2018learning, tang2023cotdet} and affordance learning from human-scene interactions~\cite{delaitre2012scene, fouhey2012people, wang2017binge}. Inspired by this ongoing discussion, we study how generative models perceive affordance by creating interactive videos.

%Donald Norman migrated this concept to human-computer interaction and later on added the concept of \textit{signifier} which emphasizes the locations of actions~\cite{norman2013design}. In his word, ``Affordances determine what actions are possible. Signifiers communicate where the action should take place. We need both.''~\cite{norman2013design}. 
\section{Preliminary: Text-to-Video Generation}
In this work, we leverage Movie Gen~\cite{meta2024moviegencastmedia} as our base text-to-video model. Due to resource limitations. we conduct our experiments on a 4B-parameter model that generates 128-frame 256p videos as a proof of concept, instead of training the official 30B-parameter model that operates at 1080p. We highlight key architectural and training aspects incorporated into our experiments in this following section. Refer to the supplementary material for more details. 

\label{sec:preliminary}
% This paragraph will be updated following moviegen!!
% Diffusion models are a class of generative models which learn to generate data by iteratively denoising samples drawn from a noise distribution. 
% Gaussian diffusion models assume a forward noising process which gradually applies noise ($\boldsymbol{\epsilon}$) to real data ($\boldsymbol{x_0} \sim p_{\text{data}}$). Concretely, 
% \begin{equation}
% \boldsymbol{x_t} = \sqrt{\gamma(t)} \ \boldsymbol{x_0} + \sqrt{1 - \gamma(t)} \ \boldsymbol{\epsilon},
% \end{equation}
% where $\boldsymbol{\epsilon} \sim \mathcal{N}(\boldsymbol{0}, \boldsymbol{I}), t \in \left[0, 1\right]$, and $\gamma(t)$ is a monotonically decreasing function (noise schedule) from $1$ to $0$. Diffusion models are trained to learn the reverse process that inverts the forward corruptions:
% \begin{equation}
% \mathbb{E}_{x \sim p_{\text{data}}, t \sim \mathcal{U}(0, 1), \boldsymbol{\epsilon} \sim \mathcal{N}(\boldsymbol{0}, \boldsymbol{I})} \left[ \left\| \boldsymbol{y} - f_{\theta}(\boldsymbol{x_t}; \boldsymbol{c}, t) \right\|^2 \right],
% \end{equation}
% where $f_{\theta}$ is the denoiser model parameterized by a neural network, $\boldsymbol{c}$ is conditioning information e.g., class labels or text prompts, and the target $\boldsymbol{y}$ can be random noise $\boldsymbol{\epsilon}$, denoised input $\boldsymbol{x_0}$ or $\boldsymbol{v} = \sqrt{1 - \gamma(t)} \ \boldsymbol{\epsilon} - \sqrt{\gamma(t)} \ \boldsymbol{x_0}$. 
\noindent \textbf{Temporal autoencoder. } %Handling high-resolution images and videos in raw pixel space is computationally expensive. To mitigate this, o
Our model encodes RGB videos and images into a learned spatiotemporally compressed latent space using a Temporal Autoencoder (TAE) and generates videos in this space. The TAE encoder is designed by inflating the image autoencoders in ~\cite{rombach2021highresolution}, adding an 1D temporal convolution after each 2D spatial convolution and an 1D temporal attention after each spatial attention. 

% Specifically, the TAE in the 4B model compresses spatially by $8\times$ and temporally by $4\times$.

\noindent \textbf{Video generation backbone. }
The model generates videos within a learned latent space as defined by the TAEs. The latent video code is segmented into patches via a 3D convolutional layer \citep{dosovitskiy2020image}, then flattened into a 1D sequence as input to the generation backbone. The generation backbone consists of Transformer~\cite{vaswani2017attention} blocks with cross-attention modules inserted between self-attention and feed-forward networks, enabling text conditioning via text prompt embeddings. The model employs UL2~\cite{tay2023ul2}, ByT5~\cite{xue2022byt5}, and Long-prompt MetaCLIP~\cite{xu2023metaclip} as text encoders, enabling both semantic- and character-level text understanding.

\noindent \textbf{Flow matching.} The model is trained with Flow Matching~\cite{lipman2023flowmatching, boffi2024flowmapmatching}, which iteratively transforms a prior Gaussian distribution into a sample from the target data distribution. During inference, an ordinary differential equation (ODE) solver transforms random noise into video latents. We use this training and inference framework for all experiments.

\section{Affordance-Aware Video Generation}
\label{sec:methods}
\begin{figure*}[!ht]
    \centering
    \includegraphics[width=\linewidth]{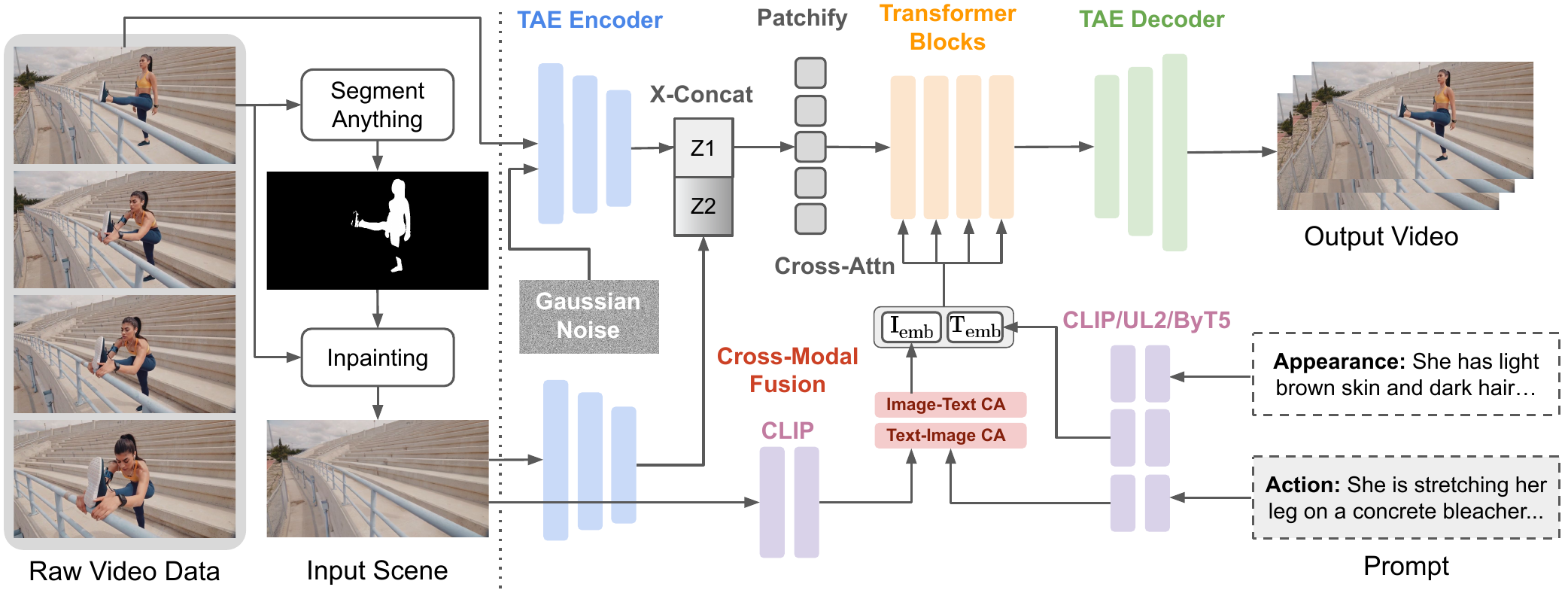}
    \caption{We start by removing humans from raw frames to create synthetic empty-scene and human-video data pairs. We employ a dual-conditioning mechanism, using channel concatenation and cross-attention, to condition the T2V model on an additional scene image. We design a fusion module to facilitate interactions between image and action-text features while locating the desired action position. The fine-tuning pipeline trains a Transformer architecture based on flow matching framework. }
    \label{fig:pipeline}
\end{figure*}

Our full pipeline is illustrated in Fig.~\ref{fig:pipeline}. We define the problem, explain data processing, and model architecture below.

%In this section, we present more details of our data preparation and model training framework. We define and clarify our task in Sec.~\ref{sec:def}, explain the basics of our base transformer-based text-to-video model in~\ref{sec:preliminary}, present our model architectural design in Sec.~\ref{sec:condition}, and show the data processing pipeline in Sec.~\ref{sec:data}. An overview of our method is shown in Fig.~\ref{fig:pipeline}

%\subsection{Preliminary}

\subsection{Task Definition}
\label{sec:def}
Let $I$ be an image of a static scene, and let $T_h$ and $T_a$ be text prompts describing a human's appearance and action. We generate a video $V$ that depicts the given scene $I$ with an inserted human matching the appearance described by $T_h$ and performing the action in $T_a$. During training and inference, we provide no explicit guidance for the human's position or pose in the scene, allowing the generative model full freedom to position the action, simulate body movements, and render the video. Note that this is not image animation; the scene image serves only as a reference for the background appearance and the presence of semantically meaningful objects. We do not require the image to appear as a frame in the video, nor do we treat the scene as a static background for pasting the human without environmental animations or camera viewpoint changes. %Our model implicitly reasons about the geometric structure, semantic layouts, and appearance features like lighting and shadow properties to harmonize a natural human insertion.

\subsection{Training Data}
\label{sec:data}
\begin{figure}[!ht]
    \centering
    \includegraphics[width=0.7\linewidth]{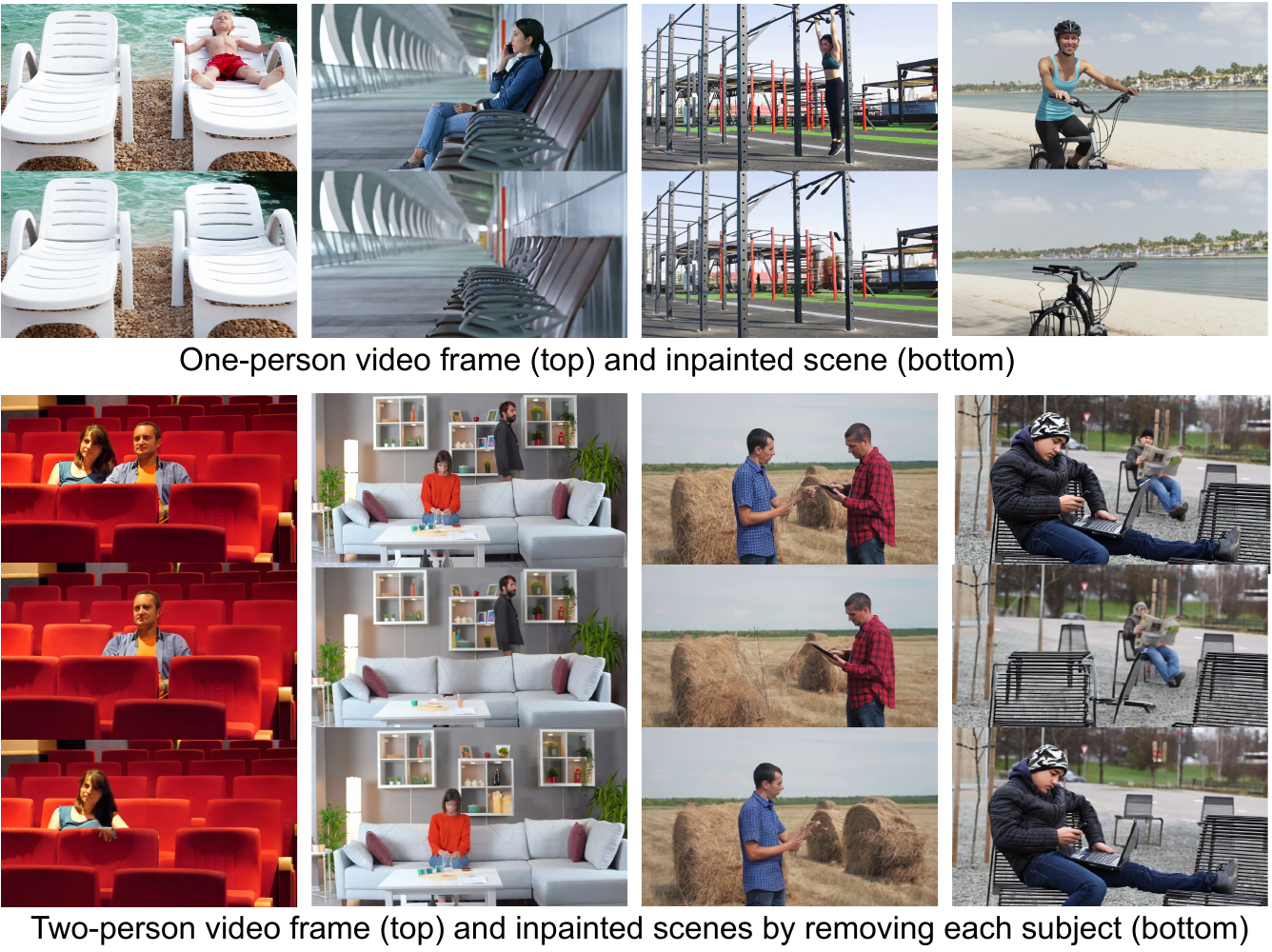}
    \caption{Representative samples of our dataset. Top row shows single-person data, while the bottom row shows double-person data. Within each row, the top figure presents the raw first frame of the video, while the bottom figure(s) show the result after detecting and removing humans from the scene. The background remains unchanged while the subject is removed. 
    }
    \label{fig:train-data}
\end{figure}

In this section, we explain our full data processing pipeline. Representative data samples are shown in Fig.~\ref{fig:train-data}. 

\noindent \textbf{Human filtering.} We curate our dataset by selecting human-related videos from the ShutterStock~\cite{sstk} text-video dataset. We apply human detection to each middle video frame and retain only those with one or two detected persons. This filtering leaves us with around 250,000 videos with one person, and another approximately 171,000 videos with two people. 

\noindent \textbf{Full body filtering.} We apply OpenPose~\cite{cao2019openpose} to videos that pass the previous stage, retaining those where knees' keypoints are visible or face's height and width fall below a threshold to avoid half-body or close-up shots.

\noindent \textbf{Pure background filtering.} We compute the color variance of background pixels in the middle frame of each video, retaining only those exceeding a threshold of 200. We also scan video captions and exclude those containing keywords like ``a pure green background.'' This helps eliminate studio-recorded videos that lack background interaction.

\noindent \textbf{Human removal.} We take the first and last frames of each video, with GroundingDINO~\cite{liu2023grounding} detecting the central human subject and language-guided SAM~\cite{kirillov2023sam} segmenting the human mask. We dilate the mask by 50 pixels to fully cover the human region and apply a text-to-image inpainting model with the negative prompt "human" for removal. For two-person videos, we remove one person at a time, creating two data samples from a single video. This results in a training dataset of \texttt{(text, image, video)} tuples representing \texttt{(action, scene, interaction)}, including 217,530 samples for single-person data and 29,700 for two-person data. We handpick 300 samples per category for validation and detail the post-processing steps for the synthetic validation benchmark in Sec.~\ref{sec:dataset}.

\noindent \textbf{Prompt rewriting.} We use LLaMA 3~\cite{dubey2024llama3herdmodels} to rewrite video captions, separating out human-related prompts ($T_a$ and $T_h$) and removing sentences that pertain solely to the background. This allows the model to learn background information purely from the visual modality rather than text, promoting multimodal information fusion.

\subsection{Conditioning Mechanism}
\label{sec:condition}

During fine-tuning, we aim to keep the original structure as much as possible, while exploring conditioning strategies to unfold a text-to-video model's innate ability on perceiving affordance from a scene image. We discuss key strategies to condition the model on an additional image input.

\noindent \textbf{Masked latent concatenation. } To maintain background consistency with the given image $I$, we concatenate the image latent $Z_1$ with the noisy latent $Z_2$ along the channel dimension before feeding them into the Transformer's backbone. Since our model is not an image animation model, we allow environmental updates driven by both the action prompt $T_a$ and natural effects such as camera movements. To achieve this, we progressively add Gaussian noise to the conditional image latent $Z_2$ with a temporal scaling factor of $\gamma=0.8$, weakening control as the video progresses until the last frame is fully masked. This decay-based control strategy ensures the scene initially matches the given image while allowing camera movement and human interaction to modify scene elements over time.

\noindent \textbf{Fused text-image feature enhancer.} We augment the cross-attention conditioning branch with a fusion module that practices mutual attention on embeddings of image and action text, drawing inspiration from~\cite{liu2023grounding, li2021grounded}. Following the original model, we concatenate three types of text embeddings (ByT5, UL2, MetaCLIP) to form a unified textual representation and use the CLIP image feature map before pooling it into a spatial-aware image embedding. We apply deformable self-attention~\cite{xia2023dat} to enhance image features and standard self-attention for text features. To promote cross-modal alignment, we introduce separate image-to-text and text-to-image cross-attention layers for feature fusion. This fusion module enhances the text-to-video model’s grounding ability, allowing it to 'locate' corresponding action regions within the image. We concatenate the fused image embedding with raw textual embeddings and input them into each Transformer block in the text-to-video model via cross-attention, as in Sec.~\ref{sec:preliminary}.

\noindent \textbf{Controlled guidance scale. } Following the practice of InstructPix2Pix~\cite{brooks2022instructpix2pix}, we leverage a controlled multi-scale guidance mechanism to control the strength of background scene image and action prompt.
%\begin{equation}
%\begin{aligned}
%\hat{e}_\theta\left(z_t, c_I, c_T\right)&= e_\theta\left(z_t, \varnothing, \varnothing\right) \\
%& +s_I \cdot\left(e_\theta\left(z_t, c_I, \varnothing\right)-e_\theta\left(z_t, \varnothing, \varnothing\right)\right) \\
%& +s_T \cdot\left(e_\theta\left(z_t, c_I, c_T\right)-e_\theta\left(z_t, c_I, \varnothing\right)\right)
%\end{aligned}
%\end{equation}
%where $c_I, c_T$ are the scene and action conditions, and $s_I, s_T$ control their strength, respectively. 
A higher image strength preserves scene consistency, while a higher text strength emphasizes human action and promotes plausible environmental updates. Training with dummy condition images helps maintain the pre-trained model's text-to-video capability and prevents overfitting to a specific dataset domain.

\subsection{Implementation Details}
We use the base text-to-video model Movie Gen~\cite{meta2024moviegencastmedia} with 4B parameters, as described in Sec.~\ref{sec:preliminary}. We train with landscape 256p, 16 frames per second, eight seconds per video. We fine-tune the full model with the text encoders frozen. We use a per GPU batch size of 1, and a learning rate of 1e-5. The training takes two days on 32 H100 GPUs.

\section{Unveiling Implicit Affordance Capability}
\label{sec:analysis}

We comprehensively analyze the implicit affordance modeling capabilities of our proposed model. In Sec.~\ref{sec:t2v} we justify that affordance perceiving information can be unveiled by investigating the the cross-attention modules, specifically which processes and regulates the CLIP text conditions. In Sec.~\ref{sec:pad} we apply our model on a real-world affordance prediction dataset. As a preliminary, the primary objective of cross-attention is to select appropriate values $\mathbf{V}$ using the attention scores $\mathbf{S}$ determined by 
$$\mathbf{S} = \operatorname{softmax}(\mathbf{Q} \mathbf{K}^T / \sqrt{d}) \in \mathbb{R}^{n \times m}$$ 
Here, $\mathbf{Q} \in \mathbb{R}^{n \times d}$ represents the projected and flattened intermediate diffusion features. $\mathbf{K} \in \mathbb{R}^{m \times d}$ and $\mathbf{V} \in \mathbb{R}^{m \times d}$ are the projected features of the input text embedding. The attention map $\mathbf{S}$ provides a physical interpretation where each entry $(i, j)$ indicates the saliency of interaction between a spatial location $i$ and a token $j$ in the prompt. This saliency reflects how strongly a particular spatial feature is influenced by or associated with a specific word, guiding the model in generating contextually relevant outputs.

\subsection{Predicting Affordances via Cross-Attention}
% Compare with pre-trained T2V
\label{sec:t2v}

We explore the implicit affordance reasoning capability of video modela by visualizing the $j$-th entry of the attention map $\mathbf{S}$ where the $j$-th token corresponds to an action-related term in the prompt. For example, given the input prompt ``a woman holding the rope and riding a horse'', we focus on visualizing the attention heatmap associated with the verb ``holding'' and ``riding''. 

\begin{figure}[!th]
    \centering
    \includegraphics[width=0.7\linewidth]{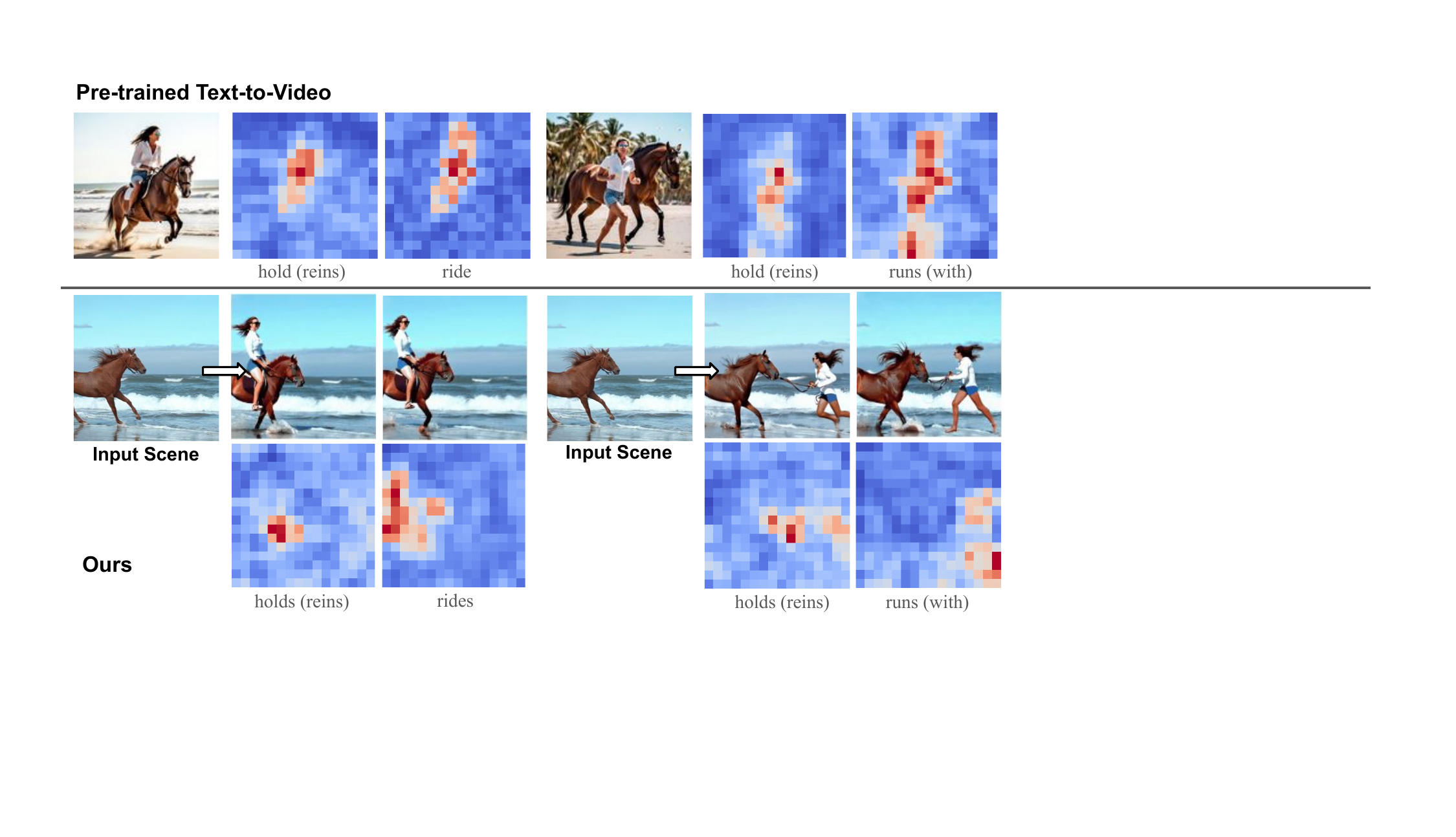}
    \caption{Cross-attention maps of the video models. Top half is the pre-trained model where the presented scene is generated by the model, and bottom half is our scene-conditioned model with a real image as input. Attention is averaged across timesteps.}
    \label{fig:attention}
    \vspace{-5mm}
\end{figure}

The top half of Fig.~\ref{fig:attention} shows the attention scores of the pre-trained T2V model. Trained exclusively on text-video pairs, the model exhibits a reasonable ability to perceive affordances while generating high-quality, faithful content. The heatmaps align well with action regions, highlighting the model’s ability to associate generated spatial features with actions. However, this correlation appears to be a by-product of video model training, as the heatmaps are conceptually intermediate steps in \textit{synthetic} video generation.

Building on this observation, we propose that conditioning the model on an additional scene enables it to perceive affordances in a \textit{given, real} image. The bottom half of the figure shows that the model accurately identifies action locations in input images and the specific environmental elements involved in the interaction. Our heatmaps reveal internal affordance knowledge, capturing interaction opportunities in real images rather than merely serving as by-products of synthetic content generation.

% In Fig.~\ref{fig:attention} we visualize the interaction between text and pixels by plotting the heatmap of cross-attention layers in the main text-to-video model. We shows how the heatmap is able to locate the correct location that each action happens as well as the intended environmental element to interact with.

\subsection{Real-World Affordance Prediction Experiment}
\label{sec:pad}
% On affordance dataset
We subsequently analyze our model’s affordance perception using classical 2D affordance detection datasets. We filter the Purpose-Driven Affordance (PAD) dataset~\cite{luo2021one}, retaining only images with no person and action verb-object pairs representing human actions (e.g., push, hit) and discarding passive object verbs (e.g., contain). This leaves us with 24 action verb categories, totaling 235 images with corresponding affordance masks. We create the prompts based on the affordance verb with LLaMA~\cite{dubey2024llama3herdmodels}, and pass in the image and prompts as inputs for our model.

\begin{figure}[!ht]
    \centering
    \includegraphics[width=0.7\linewidth]{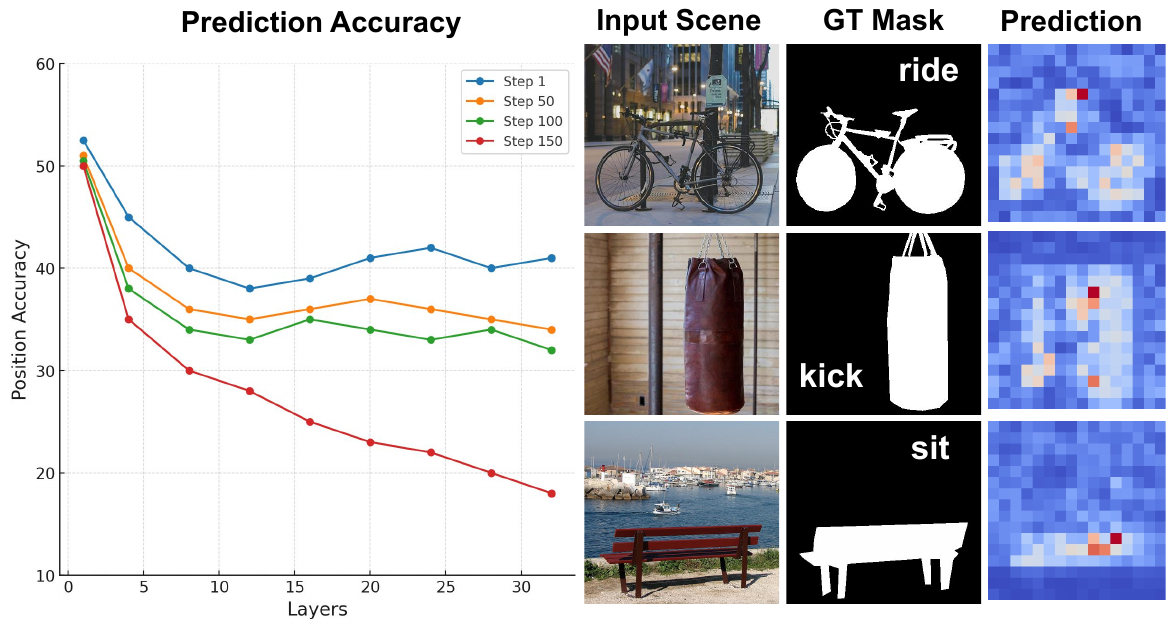}
    \caption{Affordance position accuracy across different steps and layers on a subset of PAD~\cite{luo2021one}. The attention scores indicate strong predictive ability, and visualizations show that our model accurately locates detailed affordance information.}
    \label{fig:pad}
    \vspace{-5mm}
\end{figure}

In Fig.\ref{fig:pad}, we present heatmap visualizations, derived similarly to those in Sec.\ref{sec:t2v}. We also compute the spatial accuracy (defined as pixel-wise IoU) between the binarized attention map and the ground-truth affordance mask across different layers and diffusion inference steps. We observe slightly higher scores in the initial layers, likely because the model processes semantic information early in generation. Even in the early steps, our model consistently predicts affordance through attention features. Accuracy decreases in later steps as the model shifts from perceiving high-level semantics to refining details of generated content. Peaks in the attention heatmap gradually transition from interaction regions to human content. The spatial alignment of heatmap and ground-truth affordance maps highlights the video model's ability to perceive affordance in alignment with real-world data. Our model even outperforms the ground-truth by predicting object parts relevant to each action rather than entire objects. For example, it identifies the seat of a bench where people sit, rather than its legs.

%Note that we choose to not compare with existing affordance prediction methods as our model is \textit{not} trained on such affordance annotations, 
\section{Results}
\label{sec:experiments}

We present quantitative and qualitative results of our proposed affordance-aware human video generation models. Our model effectively generalizes across a variety of environments and actions, producing realistic human-scene interactions that adhere to affordance principles.

%We present a thorough and detailed evaluation of our models. In Sec.~\ref{sec:dataset} we explain more details on preparing benchmark for evaluation. In Sec.~\ref{sec:baseline} we discuss baseline methods and alternative model designs to compare with. In Sec.~\ref{sec:qualitative} we present qualitative results of additional application scenarios, including multiple prompts for one scene, and multi-person interaction. In Sec.~\ref{sec:quantitative} and~\ref{sec:human} we discuss quantitative metrics and human evaluation protocols. 

\subsection{Evaluation Dataset}
\label{sec:dataset}
% \noindent \textbf{Training data.} We train our model on a filtered, clean subset of Shutterstock videos. The single-person dataset contains 210k tuples of \texttt{(text, scene image, human video)} with empty background environment and one person in the video. The two-person dataset contains 80k tuples of \texttt{(text, scene image, human video)} with one existing person as part of the background environment and two interactive person in the video. During training, we give these two subsets equal weight and randomly shuffle them to train together. We set aside a randomly selected subset of 1000 single and 1000 double person data samples for validation, and exclude them in the training. 

\begin{figure}[!ht]
    \centering
    \includegraphics[width=0.6\linewidth]{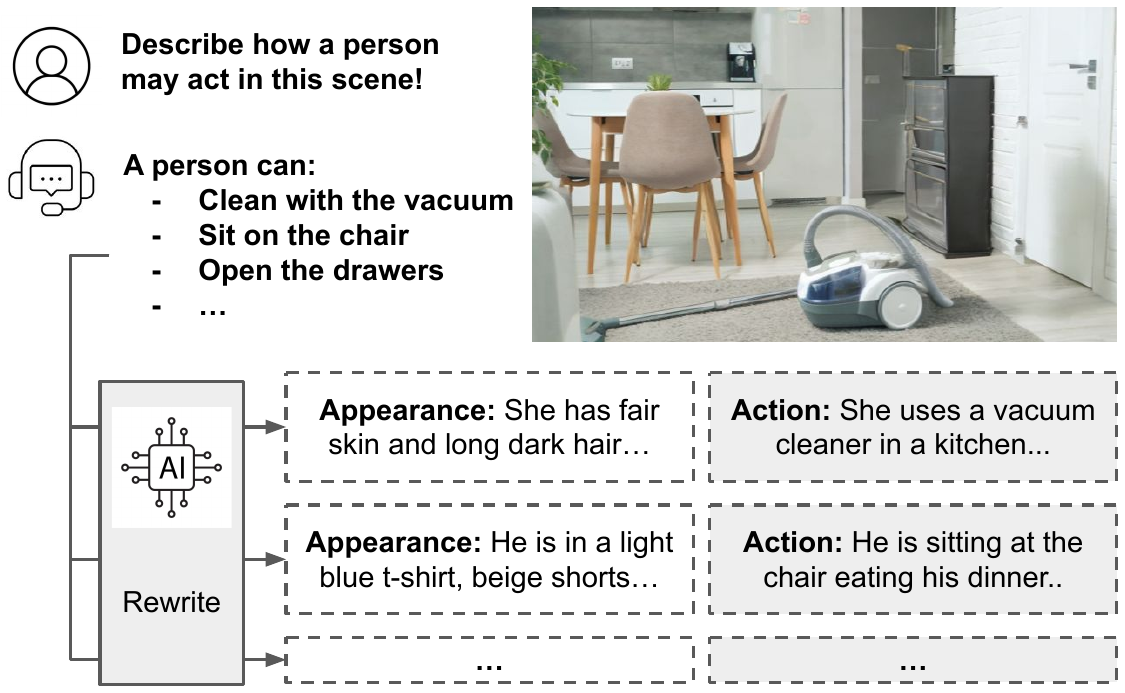}
    \caption{The synthetic action descriptions generated through our prompting process. We use a vision language AI agent to decide palusible actions in a scene, and rewrite the action into prompts. %Refer to the supplementary material for detailed prompts used and generation formats. 
    }
    \label{fig:eval-data}
\end{figure}

We aim to generate \textit{diverse} actions interacting with more than one parts of the environment, even within a fixed scene. To address this, we curate synthetic prompt sets based on real scene images. Specifically, we use a pre-trained vision-language model to generate two prompts per scene by asking, "What might a person do in this scene?" 
%We manually select two high-quality prompts from the five and have an agent rewrite them with richer details on appearance and motion.
This process yields an evaluation set of 300 images, each paired with one original and two synthetic prompts. These prompts emphasize different objects or positions within a complex environment, allowing us to assess whether our model's generative ability extends beyond central, salient objects. We repeat this process for two-person scenarios, prompting interactions with both the scene and the existing person. %We treat the existing human in the scene as part of the environment for the inserted subject to interact with. We report all results on this evaluation subset. 
Fig.~\ref{fig:eval-data} illustrates our benchmark pipeline. %We will release this evaluation benchmark for follow-up comparisons. 

% , and will release it as a benchmark for future comparison. 

\begin{figure*}[!ht]
    \centering
    \includegraphics[width=\linewidth]{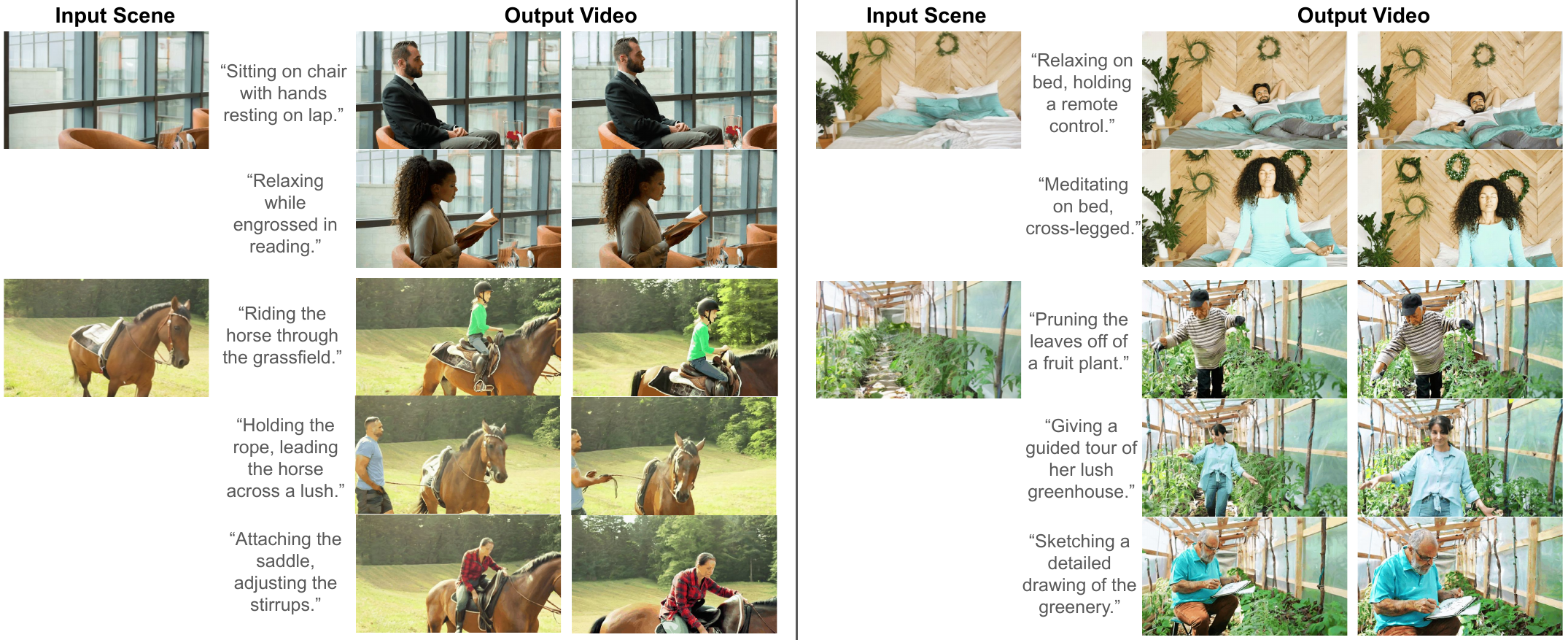}
    \caption{Our model generates diverse videos with multiple action prompts given the same scene. It identifies the correct way for an inserted subject to interact with the scene, and infers location, pose, action, spatial relationship without a pre-defined human mask prior.}
    \label{fig:diverse}
\end{figure*}

\begin{figure*}[!ht]
    \centering
    \includegraphics[width=\linewidth]{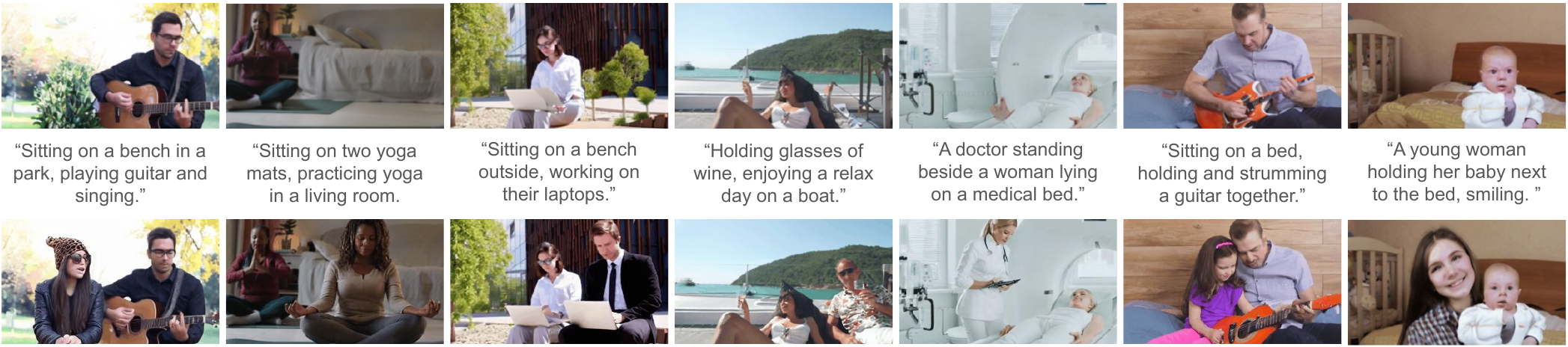}
    \caption{Our model is able to add an extra subject to a scene that contains one person. Here we consider the existing person as an organic part of the environment, and are able to synthesize interactions respecting both the background and the human in the scene. Top row is input scene image, middle row is the action prompt, and the bottom row is middle frame of the generated video. }
    \label{fig:double-human}
    
\end{figure*}

\begin{figure*}[!ht]
    \centering
    \includegraphics[width=\linewidth]{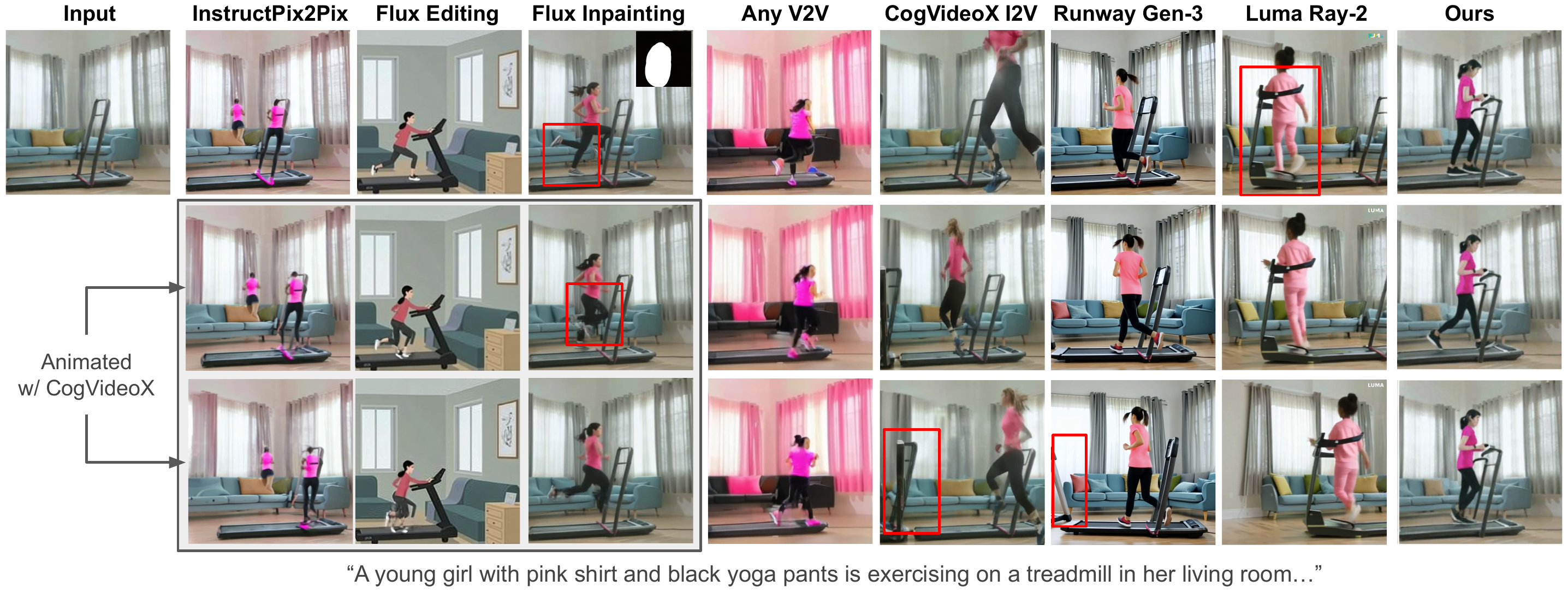}
    \caption{Comparison with baseline methods. Three rows are the first, middle and last frames for each method. The left three columns' models edit a static frame and  animate it. The next four edit video directly. Note that Flux~\cite{flux2024} Inpainting  requires a user-defined mask as input, which eases the task and greatly assists the model in predicting human position. Yet, our model outperforms baselines in terms of human placing, motion simulation and appearance rendering. See video results and more comparison in the supplementary materials.}
    \label{fig:baseline}
\end{figure*}

%\begin{figure}[!ht]
%    \centering
%    \includegraphics[width=\linewidth]{figures/ablation.pdf}
%    \caption{Compare our model with alternative designs justify our conditioning mechanism, which is able to both preserve the scene details and generate plausible, high-quality interaction. }
%    \label{fig:ablation}
%    \vspace{-3mm}
%\end{figure}

\subsection{Baselines and Ablation}
\label{sec:baseline}

\noindent \textbf{Baselines.}
To the best of our knowledge, there is no existing work on generating human videos in a scene without location or pose control. We therefore compare our methods with generic image/video editing and image-to-video solutions not tailored for humans. We compare with three image-based models: (1) \textbf{InstructPix2Pix~\cite{brooks2022instructpix2pix}} where we directly apply an image editing model on the empty scene image with the prompts. (2) \textbf{Flux Editing} which trains instructional image editing on Flux~\cite{flux2024}. (3) \textbf{Flux Inpainting} where we provide a groundtruth human mask as the inpainting position. We then compare with instruction-based video editing method (4) \textbf{AnyV2V~\cite{ku2024anyv2v}} where the scene is repeated for 2 seconds to a video, and then edited based on a prompt. We additionally compare with one open-source and two commercial video generation models (5) \textbf{CogVideoX~\cite{yang2024cogvideox}}, (6) \textbf{Runway Gen-3~\cite{runway2024gen3}} and (7) \textbf{Luma AI Ray-2~\cite{luma2024ray2}} where we apply image-to-video on the scene with a prompt. For (1), (2), (3), we attach a CogVideoX image animation model to the image results, exploring their potential of generating interaction videos in the same setting as ours. Note that we only do small scale visual comparison and human evaluation on (6) and (7) without quantitative metrics as they do not have a free API available.

\noindent \textbf{Ablation studies.} We compare with alternative designs of our model that remove key features including latent concatenation, fused cross-attention, and Gaussian noise decay. Due to space limits, parts of the ablation results are presented in the supplementary material.

\subsection{Qualitative Evaluation}
\label{sec:qualitative}
\noindent \textbf{Human-scene interaction. } Fig.~\ref{fig:teaser} presents inserting a human into a scene based on an action prompt. The model maintains pixel-level scene consistency while placing the subject correctly without a predefined mask. The generated video features natural camera movements, object updates in response to human actions, and scene animations.

\noindent \textbf{Diverse affordance. } In scenes with complex layouts and multiple interaction possibilities, our model inserts subjects while accounting for diverse scene elements and action-affording objects. Fig.~\ref{fig:diverse} illustrates how our model determines subject placement and imagines body poses for different actions (e.g., riding vs. standing beside a horse).

\noindent \textbf{Human-human interaction. } Fig.~\ref{fig:double-human} shows adding a  subject to interact with both the scene and an existing person who is considered a part of the scene. %Video shows natural interaction between the given and added subjects.

\noindent \textbf{Baseline comparison. } 
Fig.~\ref{fig:baseline} demonstrates that our model achieves the highest semantic alignment and visual fidelity. Instruction-based editing methods like InstructPix2Pix and AnyV2V~\cite{brooks2022instructpix2pix,ku2024anyv2v} generate distorted human bodies and misattribute prompt concepts (e.g., applying ``pink'' to the treadmill or curtain instead of clothing). Editing methods based on better image model Flux~\cite{flux2024} does not preserve scene styles and generates cartoon videos. Flux~\cite{flux2024} Inpainting distorts human bodies even when provided with an additional mask and fails to preserve pixel details in masked background regions (the yellow pillow disappears). Current best open-sourced and commercial image-to-video models like CogvideoX\cite{yang2024cogvideox}, Runway Gen-3~\cite{runway2024gen3} and Luma Ray-2~\cite{luma2024ray2} all misinterpret the treadmill’s affordance, place subjects in the wrong direction, and hallucinate another treadmill on the left. Our models stand out by successfully preserving the background and simulating natural interactions between the subject and the treadmill.

% \noindent \textbf{Ablation comparison. } As shown in Fig.~\ref{fig:ablation}, our dual stream conditioning approach with both latent concatenation and feature enhanced cross-attention proves to be the best way of conditioning a T2V model on the scene image. Without latent concatenation, the model generates something semantically similar but not pixel-wise the same. Without fused cross-attention modules, the model is prone to generating distorted, unreasonable motions. 

% A few generated results
% same scene, different prompts. same prompt, different scenes.
% scene in the validation set + new prompt, new scene + new prompt
% a figure comparing with baseline. 

\subsection{Quantitative Evaluation}
\label{sec:quantitative}

We evaluate our model based on human video faithfulness, text-video alignment, and action quality. This corresponds to three major quantitative metrics: (i) \textbf{FVD (Fréchet Video Distance)}~\cite{unterthiner2018fvd}, which quantifies the similarity between real and synthetic video embedding distributions. (ii) \textbf{CLIP}~\cite{radford2021clip} similarity, which computes the average embedding similarity between the input prompt and each generated frame to assess prompt alignment. (iii) \textbf{Action Score}, computed by querying a pre-trained VQA model~\cite{zhang2024llavanext-video} with ``What action is the person performing in this video?'' and measuring the CLIP similarity between the recognized motion and the ground-truth action prompt. The Action Score helps isolate interaction accuracy by reducing the influence of appearance. For image-only baselines, we compute metrics on the animated video sequence using CogVideoX~\cite{yang2024cogvideox}, one of the best open-sourced image animation models. 

We quantitatively compare our model with baselines and ablated variants. Results in Tab.~\ref{tab:baseline} show that our model consistently outperforms others in human video quality, text alignment, and action faithfulness.

\begin{table}[t]
\centering
\begin{minipage}[t]{0.48\linewidth}
\centering
\caption{Quantitative evaluation shows our method consistently outperforms baselines and ablation methods.}
\label{tab:baseline}
\begin{tabular}{lccc}
\toprule
Model & CLIP $\uparrow$ & FVD $\downarrow$ & Action $\uparrow$ \\ \midrule
InstructPix2Pix   & 0.19 & 302 & 0.14 \\
Flux Inpainting   & 0.40 & 174 & 0.65 \\
Flux Editing      & 0.23 & 332 & 0.63 \\
AnyV2V            & 0.23 & 290 & 0.33 \\ 
CogVideoX         & 0.38 & 199 & 0.69 \\
\midrule
w/o x-concat      & 0.46 & 185 & 0.76 \\
w/o cross-attn    & 0.59 & 220 & 0.55 \\
w/o fusion        & 0.65 & 171 & 0.85 \\
Ours              & \textbf{0.67} & \textbf{168} & \textbf{0.88} \\
\bottomrule
\end{tabular}
\end{minipage}%
\hfill
\begin{minipage}[t]{0.48\linewidth}
\centering
\caption{Human evaluation preference comparison with baseline approaches.}
\label{tab:baseline-he}
\setlength{\tabcolsep}{4pt} % reduce horizontal padding
\begin{tabular}{lcccc}
\toprule
Model & SC (\%) & HQ (\%) & PA (\%) & AP (\%) \\ \midrule
InstructPix2Pix   & 100 & 98  & 100 & 96  \\
Flux Editing      & 87  & 94  & 99  & 97  \\
Flux Inpainting   & 95  & 79  & 60  & 57  \\ 
AnyV2V            & 100 & 100 & 100 & 98  \\ 
CogVideoX         & 68  & 87  & 74  & 89  \\
Runway Gen-3      & 54  & 65  & 67  & 70  \\
Luma Ray-2        & 55  & 59  & 69  & 75  \\
\midrule
w/o x-concat      & 99  & 48  & 53  & 76  \\
w/o cross-attn    & 73  & 89  & 61  & 69  \\
w/o fusion        & 54  & 52  & 58  & 60  \\
w/o noise decay   & 76  & 48  & 56  & 53  \\
\bottomrule
\end{tabular}
\end{minipage}
\end{table}

\subsection{Human Evaluation}
\label{sec:human}
We supplement our analysis with a structured A/B test human evaluation. We assess the results based on four criteria: \textbf{(i) Scene consistency (SC)} evaluates how well the video preserves the original scene, even with flexible camera angles and scene motions. \textbf{(ii) Human quality (HQ)} assesses the realism of the generated human body. \textbf{(iii) Text-prompt alignment (PA)} evaluates how accurately the generated action and appearance match the given prompt. \textbf{(iv) Affordance prediction (AP)} assesses the interaction plausibility between the subject and the scene. Tab.~\ref{tab:baseline-he} presents the percentage of subjects preferring each model, demonstrating that our model is consistently perceived as more realistic, natural, and capable of producing reasonable interactions compared to baselines and ablations.  % Refer to the appendix for more details on human evaluation setup.

% We recruit 20 human users as the evaluation subjects. They are randomly presented the scene image, text prompt, and a pair of results from our full model and one baseline model, and are asked to choose a better one. Each subject is presented with 100 pairs of images, randomly chosen from 100 validation images, 3 prompts per image, and 3 baseline models. In Tab. we present the percentage of preference of our model results versus the baseline versions.

\section{Conclusion}

We explore the ability of text-to-video models to perceive affordance and reason about interaction through the task of populating empty scenes with moving humans. Beyond a creative application, we show that video generative models implicitly learn affordance and can simulate affordance-aware activities through extensive analysis of attention features. We provide preliminary insights into effectively leveraging video generative models beyond appearance rendering toward interaction simulation.

\bibliographystyle{assets/plainnat}
\bibliography{paper}

\end{document}